\documentclass[10pt, a4paper]{article}
\usepackage{lrec2022} 
\usepackage{multibib}
\newcites{languageresource}{Language Resources}
\usepackage{graphicx}
\usepackage{tabularx}
\usepackage{soul}
\usepackage{todonotes}
\usepackage{titlesec}
\titleformat{\section}{\normalfont\large\bfseries\center}{\thesection.}{1em}{}
\titleformat{\subsection}{\normalfont\SmallTitleFont\bfseries\raggedright}{\thesubsection.}{1em}{}
\titleformat{\subsubsection}{\normalfont\normalsize\bfseries\raggedright}{\thesubsubsection.}{1em}{}
\renewcommand\thesection{\arabic{section}}
\renewcommand\thesubsection{\thesection.\arabic{subsection}}
\renewcommand\thesubsubsection{\thesubsection.\arabic{subsubsection}}

\usepackage{epstopdf}
\usepackage[T1]{fontenc}
\usepackage[utf8]{inputenc}
\usepackage{hyperref}
\usepackage{xstring}

\usepackage{color}

\title{\textbf{Language technology practitioners as language managers:\\ arbitrating data bias and predictive bias in ASR}}

\name{Nina Markl, Stephen Joseph McNulty} 

\address{University of Edinburgh \\
         nina.markl@ed.ac.uk, stephen.mcnulty@ed.ac.uk\\}

\abstract{
Despite the fact that variation is a fundamental characteristic of natural language, automatic speech recognition systems perform systematically worse on non-standardised and marginalised language varieties. In this paper we use the lens of language policy to analyse how current practices in training and testing ASR systems in industry lead to the data bias giving rise to these systematic error differences. We believe that this is a useful perspective for speech and language technology practitioners to understand the origins and harms of algorithmic bias, and how they can mitigate it. We also propose a re-framing of language resources as (public) infrastructure which should not solely be designed for markets, but for, and with meaningful cooperation of, speech communities. \\ \newline  }

\begin{document}

\maketitleabstract

\section{Introduction}

All language communities, even monolingual ones, show linguistic variation.  The co-existence of multiple different ways of communicating the same meaning is a fundamental characteristic of natural language. Speakers can employ different words to refer to the same object (e.g. ``film'' and ``movie''), pronounce the same word differently (e.g. ``data'' and ``data''), address interlocutors differentially depending on the context (e.g. ``you'', ``yous'', ``y'all'' in many varieties of English), and even utilise different sentence structures (e.g. ``data is'' and ``data are''). Despite that fact that this variation is inevitable, people still form judgements about them. Very often, these judgements reflect biases about (groups of) people, not language per se. 

These social-linguistic judgements contributes to differential access to, and performance of, language technologies for speakers of the over 7000 language varieties\footnote{``Language variety'' refers to languages (e.g. English), ``dialects'' (e.g. Scottish English) and accents (e.g. Standard Scottish English). The linguistic features characterising a variety are called ``variants''.} spoken in the world. Most language communities globally do not have access to them at all, and within those that do, performance for speakers of non-standard(ised) and marginal(ised) varieties is worse. For automatic speech recognition (ASR) systems, this “predictive bias”, defined by \newcite{shah-etal-2020-predictive} as a systematic error disparity between different user groups, arises in part from data bias in the speech datasets used to train and test them. In this paper, we use the lens of ``language policy'' to understand the origins and consequences of this data bias, and to facilitate its mitigation. We contend that, perhaps unknowingly, organisations – and particularly individuals – involved in the design and creation of these datasets, whether crowdsourced or curated, perform the function of ``language policy arbiters''. In their selection of widely spoken, prestigious (and often commercially-viable) varieties, these individuals effectively marginalise speakers of minority or lesser-used languages or forms of language. This marginalisation may take the form of limiting access to these technologies and exacerbating stigma towards some varieties in their wider application, thereby amplifying systemic discrimination against particular groups and their language(s). Yet, simply by recognising the need for proactive, diversity-oriented language management – and their role in engendering it – speech and language technologists can work to mitigate such harms, and work towards more equitable and inclusive technologies. 

\section{Predictive bias in ASR}\label{ASRbias101}

Recent work shows that state-of-the-art commercial English language ASR systems display significant predictive bias for African American English (AAE) and  some regional varieties of English. \newcite{Koenecke2020} document dramatic racial error disparities for ASR systems sold by Google, Amazon, Microsoft, IBM and Apple, with much larger error rates for Black speakers of AAE than White speakers of Californian English. Overall, recent research suggests that this predictive bias is driven by under-representation of AAE in training data for both acoustic models \cite{Koenecke2020} and language models \cite{Martin2020} used by commercial ASR systems. \newcite{Koenecke2020} find error disparities based on pronunciation differences, while \newcite{Martin2020} show that Google Cloud Speech-to-Text handles AAE syntactic features such as ``habitual be''\footnote{A common feature of AAE not found in Mainstream US English e.g. “I be in my office at 7.30” which is equivalent to MUSE “I am usually in my office at 7.30” \cite{green_2002}.} poorly. A slightly older set of studies has shown similar error disparities for different regional varieties of English, including Scottish English and Southern U.S. English in products sold by Google and Bing \cite{Tatman2017,Tatman2017a}. Notably, this apparent data bias is not limited to commercial ASR systems, as Mozilla’s open-source DeepSpeech system trained on their crowdsourced CommonVoice corpus also performs significantly worse for AAE and Indian English than Mainstream US English \cite{Martin2020,Meyer2020}. Other work has focused on the use of ASR as an assistive technology and found that most major systems perform poorly for Deaf and hard of hearing \cite{Glasser2019}, and dysarthric users  \cite{DeRussis2019,Young2010}. 

\subsection{Harms of predictive bias}

While technical research on bias mitigation is important, especially because particular model structures can amplify data bias \cite{Hooker2021}, it is crucial to consider the socio-historical origins of predictive bias and its consequences. Most obviously, speech recognition is a component of voice user interfaces which can be used as assistive technologies to access mobile devices and computers. As (mobile) computing becomes increasingly ubiquitous, predictive bias could severely disadvantage AAE speakers and other speakers of stigmatised and under-represented language varieties in completing everyday tasks such as making phone calls, searching for information on the web and sending emails, and engaging agents in private and public sector contexts. Recently speech recognition has joined other AI technologies in moving into very high-stakes contexts such as hiring and healthcare \cite{leeNextBigTech2021}\footnote{Amazon even sells an ASR system specifically for medical transcription: https://aws.amazon.com/transcribe/medical/}. Companies like HireVue\footnote{https://www.hirevue.com/}, for example,  claim to use “voice data” including information about voice quality and lexical choice to pre-screen and rank job applicants \cite{RaghavanBarocasKleinbergLevy2020}. While HireVue has recently passed an independent audit of their algorithmic systems, according to which their training data is balanced by race, gender, region and job title, accent-based bias among first and second language speakers of English has not been studied\footnote{https://www.hirevue.com/blog/hiring/industry-leadership-new-audit-results-and-decision-on-visual-analysis}. HireVue and its competitors also offer customisation of training data for client companies, which makes identification and mitigation of data bias in practice particularly difficult \cite{RaghavanBarocasKleinbergLevy2020}. By disadvantaging marginalised speech communities in accessing technology and resources (up to and including employment), predictive bias can further reify and entrench existing linguistic, and by extension, social hierarchies.

\section{Language Policy}\label{sec:socio101}

As \newcite{blodgett-etal-2020-language} show, discussions of “bias” in the language technology literature often lack grounding in the broader socio-historical context of users or the system, failing to spell out what exactly is meant by “bias,” who is harmed by it and how it relates to larger power structures. In this paper, we use the sociolinguistic concept of “language policy” to understand both where data bias comes from and whom it harms. \par 

Language policy relates to the rules, conventions, choices, values, ideas, or discourses which govern the way that we use or think about languages and their speakers \cite{Spolsky2004,Johnson2013}. These policies can be either explicit or overt, as is the case of language legislation or institutional language policy documents, or can be concealed, covert or de facto – often couched in decisions or actions not specifically related to languages, or in implicit judgements about them or those who speak them \cite{Shohamy2006}.  For \newcite{Spolsky2004} it is composed of three distinct but interrelated phenomena, which we will explain in turn. \textit{Language practices} refer to conventionalised or patterned language behaviours; \textit{language ideologies} are value-based judgements of specific language varieties and variants, and by extension their speakers and communities; and \textit{language management} refers to attempts to modify language practice and language ideologies. 


\subsection{Language practices} 

As we have noted, language use is characterised by variation. In an influential formulation, \newcite{weinreichherzoglabov1968} refer to this variation as ``orderly heterogeneity''. That is, variation in language is patterned and rule-governed. Individuals and speech communities use this variation to construct social identities in interaction \cite{Eckert2012}. These linguistic choices, which are constrained by the social norms transmitted within a community, make up the community's language practices \cite{Spolsky2004}.\par 
Understanding patterns of language variation is crucial to identifying the sources of predictive bias in ASR (and other speech and language technologies) and developing mitigation strategies. For instance, some earlier work on predictive bias in ASR noted apparent differences according to speaker gender \cite{Adda-Decker2005,Benzeghiba2007,Tatman2017}. \newcite{Adda-Decker2005} locate the source of better performance for women's speech as compared to men's speech in data bias in training and test datasets. In the English language broadcast news training and test corpora \newcite{Adda-Decker2005} use, women are more likely to be newscasters and interviewers who adopt a formal speech style, while men are more likely interviewees whose speech is more often unplanned and conversational and thus characterised by repetitions, phonetic reduction, back-channels and filled pauses. In addition to the conversational roles of women and men this these datasets, they also attribute differences to broader gendered patterns of language use, whereby women tend to avoid stigmatised linguistic features more than men \cite{Labov1990}. Overall, the more formal speech styles associated here with women are easier to process for the ASR system\footnote{\newcite{garnerinInvestigatingImpactGender2021} show that when women's speech is under-represented in training sets of read speech, performance is significantly better for men. Notably, adding more women's voices improves performance for women without degrading performance for men.}. This gendered pattern in language use is also reflected in \newcite{Koenecke2020}, who find that commercial ASR systems are more error-prone for men. An analysis of their test set shows that men are, generally speaking, more likely to use higher rates of non-standard forms \cite{Koenecke2020}. This ``gender gap'' in ASR performance and speech patterns is furthermore substantially larger for Black speakers, highlighting that race and gender as interacting axes of oppression cannot be considered separately, as has long been noted by Black feminist scholars (e.g. \newcite{Crenshaw1991}. In addition to gender and race, other relevant social factors conditioning language variation are socio-economic class, educational background, linguistic background (in particular, whether speakers are first or second language speakers), disability and ethnicity (e.g. \newcite{herk2018what}). Which of these factors are particularly important depends on the specific sociolinguistic context. Generally, varieties spoken by powerful groups within a society or societal context (e.g. higher social class groups, White groups, particular geographical areas) become associated with prestige (due to their association with power). Often these prestigious varieties are also ``standard varieties'', codified in prescriptive (rather than descriptive) grammars and taught in the education system \cite{herk2018what}. Poor ASR performance on non-standard varieties, then, is more likely to affect already marginalised speech communities.

\subsection{Language ideology} 


From a linguistic perspective, no language variety is inherently “better” or “worse” than any other. However, because language (variation) is always situated within larger social contexts, specific ways of speaking can become indices of particular social identities. Language users create beliefs about language to explain and justify these (arbitrary) associations between speaker and form. As \newcite[37]{Irvine2000} put it, these beliefs “locate linguistic phenomena as part of, and evidence for, what [language users] believe to be systematic behavioral, aesthetic, affective and moral contrasts among the social groups indexed”. 

Language users (all of us) lean on these ideologies when we make judgements about other people (albeit often unconsciously). Like other ideologies, they often seem to reflect ``common sense'' and resisting them requires conscious effort. They also have real implications for, in particular, marginal(ised) groups. For example, many studies show that second language speakers of English are less likely to be hired \cite{Hosoda2010,Timming2016} and are frequently rated ``less credible'' \cite{LEVARI20101093} than first language speakers. There are further very strong language ideologies around ``professional'', ``educated'' and ``articulate'' speech \cite{Lippi-GreenRosina2012Ewaa,Baratta2017}. These ideologies are underpinned by broader structural biases within a society such as racism, classism and sexism. It is because of those broader structures that some social groups  (e.g. White, upper and middle class, men) have more power relative to others (e.g. Black and non-white, lower and working class, women and non-binary people), and as a result, their speech becomes associated with power and prestige. Notably, language ideologies are not applied in the same way in every context. This makes sense if we recall that the supposed attitudes about particular linguistic features aren't about language per se, but about the social identities they are associated with. For example, some voice qualities like creaky voice (``vocal fry'') are more stigmatised in young English speaking women than men \cite{AndersonRindyC2014Vfmu}\footnote{\newcite{AndersonRindyC2014Vfmu} conclude that women should avoid creaky voice. We reject this conclusion, and point instead to \newcite{chao_bursten_2021} for a detailed feminist critique of the response to women’s creaky voice(s).}.  

Language ideologies feed into speech and language technologies in many different ways. As we explore in this paper, they influence which kind of language we use to train and test language technologies, and as a result, who is most impacted by predictive bias. But speech and language technologies also reinforce language ideologies. Better performance on some varieties emphasises their status. And even in cases where there's no predictive bias, they can reinforce existing ideologies. For example, HireVue (and similar models) may not show predictive bias as such, but since they are trained on interviews with successful job applicants, language ideologies around ``professionalism'' as expected during a job interview are encoded. At first glance it may seem fairer if these harsh prejudices are part of an algorithmic system since they are at least applied equally (e.g. “vocal fry” = “bad,” “long sentences” = “articulate” = “good”).  But more privileged people are much more likely to have access to ``the right way of speaking'' in an interview, for example because they have been taught how to speak during interviews and what kind of language (features) to avoid.  According to the HireVue audit report, applicants from ``minority'' backgrounds are more likely to give very short answers which potentially puts them at a disadvantage as the system does not pose follow up questions.

Beliefs and ideologies about language varieties are inevitable and omnipresent. They simply represent ``what people
think should be done'' with regards language use within specific societal contexts \cite[14]{Spolsky2004}. It is when these sets of preconceived judgments begin to affect language-related decisions that we enter the realm of language management.

\subsection{Language management} 

Language management occurs across all spheres and sectors of society, and involves a wide and diverse range of actors \cite{Spolsky2004,Blommaertetal2009,Hornbergeretal2007}.  What is crucial to remember, is that, even when pertaining to the most mundane, mechanical and technical actions or decisions, language policies are never neutral. By its very nature, language management involves taking a stance on language varieties and variation, by deciding which forms of speech are appealing, acceptable or correct, and which are unattractive, inferior or simply ``wrong''. Moreover, as \newcite{Tollefson1991} and \newcite{Shohamy2006} note, language management often serves to create, reify and reproduce unequal power divisions within society: privileging speakers of dominant, prestigious varieties (e.g. native speakers of a standard form of English) and further marginalising people who use stigmatised forms of language (e.g. non-native speakers of [a non-standard] English, or speakers of minority languages). Moreover, as \cite{Wiley2012} highlights, the absence of an official policy or non-consideration of issues related to equality and diversity in language, often serves only to reinforce the power and hegemony of prestige varieties, and marginalise others: ``The lack of recognition of `nonstandard' varieties of language […] positions their speakers as merely `substandard' articulators of English.'' Inaction, therefore, is action. As noted previously, language managers, planners or policy actors can take many forms. According to \newcite{RN578}, however, certain individuals are endowed with a disproportionate amount of power within specific language policy processes. As a result of their position of influence within a given organisational, institutional or social hierarchy, these ``language policy arbiters'', through their interpretations and ideological reflexivity (or lack thereof), can influence how language policies are created or implemented \cite{Hornbergeretal2007}. Given the possible impacts of their actions, if social inequalities are truly to be redressed, it is essential that these individuals recognise how much power they wield. The design and creation of speech technologies, we believe, constitutes a form of language management with consequences across societal scales, and its designers and operators perform the role of language policy arbiters for their end users, as well as for society more generally. 

\section{State-of-the-art: training \& testing} 

An example of this form of language management would be the curation of speech datasets used in the training and testing of ASR systems. It is through this process that decisions about what kind of language to include or exclude in training and test datasets are made. These decisions then shape for which kinds of language, and therefore for which kinds of speakers, these technologies are useful rather than harmful.

\subsection{Training ASR in industry}

ASR systems by corporations like Amazon and Google, or large foundations such as Mozilla, are trained on very large datasets. In the case of commercial ASR these datasets consist (at least in part) of voice commands and dictation snippets which are collected from customers during their interactions with voice user interfaces and transcribed by employees\footnote{With consent of the users, as indicated in the privacy notices of e.g. Apple, Microsoft, Amazon and Google}.

Mozilla's corpora are made up of voice recordings which are submitted, transcribed and validated by volunteers via an online platform\footnote{https://commonvoice.mozilla.org/}. As explored in \ref{ASRbias101}, both types of systems exhibit predictive bias towards less prestigious varieties, in particular African American English. In the following section, we explore how corporate language policies influence the apparent data bias giving rise to these error disparities.

\subsubsection{Corporate: Proprietary user data}


Corporations like Amazon, Google and Microsoft do not provide detailed model documentation for the ASR systems they sell to third parties (e.g. Amazon Transcribe, Google Cloud Speech-to-Text) or the ones embedded in their own products such as voice user interfaces (e.g. Siri, Alexa, Cortana) and video platforms (e.g. YouTube captions) but their privacy notices and academic publications suggest that large proprietary datasets which include data collected from users are involved. For example, \newcite{Google2018} (Google) present a system which is trained on “representative voice search data” from their user base. Similarly, Facebook AI trained a multilingual ASR system on “publicly shared user videos” in 51 languages \cite{pratap2020massively}. A fundamental problem with training on user data is that even if this data is “representative” of the user base, the user base is not necessarily representative of the population at large. According to a 2021 Pew Research Center survey, 85\% of residents of the United States own a smartphone\footnote{https://www.pewresearch.org/internet/fact-sheet/mobile/}. However, there are still quite big gaps between different age and social class groups. There are further even larger gaps in home broadband access depending on income in particular. As has been raised in the context of large language models, while digital spaces are in in theory ``open to everyone'', participation in online communities is not equally accessible or attractive everyone \cite{Bender2020}. Any dataset based on online communication, then, risks mis- or under-representing marginalised (speech) communities who may not be able or willing to in participate \cite{Bender2020}. Indeed, the findings by \newcite{Koenecke2020}, \newcite{Martin2020} and \newcite{Tatman2017a} suggest that, in the context of US English, Black talkers in particular remain under-represented. To avoid predictive bias, data from different groups would have to be balanced rather than merely representative of the (skewed) population distribution \cite{SureshGuttag2021,barocas-hardt-narayanan}\footnote{As \newcite{Hooker2021}, notes, the fact that most “real-world” data have skewed distribution is why it’s important to focus on mitigating bias through model choice too.}. \par 

Big (speech and language) technology companies do not tend to have publicly available officially declared language policies. However, as alluded to above, just because there is no official document outlining a language policy, it does not mean that there is no policy in place. Some language policy scholars such as \newcite{Schiffman1996} and \newcite{Shohamy2006} distinguish between \textit{de jure} and \textit{de facto} language policies. Even in the absence of the former, de facto policies can still arise, often on the basis of what people in a particular context find to be sensible, convenient or common sense. In this context, beliefs about language (i.e. language ideologies) can be particularly influential \cite{Shohamy2006}. A key aspect of language management is the selection of a particular language variety to be used in a particular context. In the context of speech and language technologies, this selection process includes the choice of a particular variety to train and test a system on, and consequently, develop for. For example, \newcite{BenjaminRuha2019Rat:} quotes a former Apple speech technology researcher working on Apple's voice assistant Siri asking their supervisor in 2015 why AAE was not a priority while support for other varieties of English such as Singaporean English was being developed. The response: “Well, Apple products are for the premium market.” \cite[15]{BenjaminRuha2019Rat:}. This statement expresses a language ideology held by (at least a part of) the corporation: AAE is not spoken by ``the premium market'' and  AAE speakers do not (or cannot afford to) buy ``premium products''. Assuming that Apple's main goal is to attract (and keep) the ``premium market'' as is implicit in the quote above, only developing ``premium'' linguistic varieties is a good investment. This ideology is the company's de facto language policy: AAE is not supported by the company. By applying this economic reasoning to language varieties (and their speakers), Apple also reinforces existing ``linguistic markets'' \cite{Bourdieu1977}. It’s perhaps not surprising that \newcite{Koenecke2020} found the racial gap in predictive errors to be largest, and overall performance on AAE to be worst for Siri (as compared to other systems tested). More broadly, selecting language varieties based on their perceived value on the (linguistic) market means that varieties spoken by marginalised or small communities are less likely to be supported. Differences in language policy between corporations are also reflected in the different sets of languages they select. Google has the largest range of language varieties, including national varieties for languages like Arabic, Urdu, English and Spanish\footnote{https://cloud.google.com/speech-to-text/docs/languages}. While smaller national and regional languages spoken in Europe (like Macedonian and Basque) are supported, the same can only be said for languages with larger speaker populations outwith Europe like Uzbek, Zulu, Amharic, and Gujarati, highlighting a general global skew in speech technology availability. Similarly, Apple’s Siri is offered in US Spanish and two post-colonial English varieties (India \& Singapore) but does not support any languages indigenous to Africa, the Americas, Oceania or the Indian subcontinent. These choices do not just impact current and future customers of these technology corporations: Apple, Google and Microsoft sell their speech recognition services to third parties, and their choices (of data and algorithms) likely impact the way smaller companies act.

\subsubsection{Open-source: Crowdsourcing}

The most obvious alternative to this purely market-driven model of technology development already in use today are open-source and crowdsourced technologies, such as Mozilla’s DeepSpeech ASR system and CommonVoice collection of crowdsourced speech datasets\footnote{https://commonvoice.mozilla.org/en}. The latter currently covers 76 languages. Volunteers contribute by reading out sentences which are recorded via an interactive interface and validated by other volunteers. All contributors can optionally provide information about their gender, age and accent. CommonVoice does not appear to have a top-down policy for selecting language varieties. Volunteers can request the initiation of a corpus for a new language. The accent labels available for volunteers seem to be selected by community members\footnote{https://discourse.mozilla.org/t/spanish-accents/35638}, with Spanish varieties defined in geographic terms while German varieties are defined as national varieties (eliding variation within nation states). Similarly, the English corpus contains “Scottish English” and “England English” alongside a very broad “US English,” making comparisons of sampling bias very difficult. Mozilla is currently in the process of replacing this apparent ``null policy'' with a declared ``languages and accent strategy''\footnote{https://discourse.mozilla.org/t/common-voice-languages-and-accent-strategy-v5/56555}. This new policy has at least in part been crowdsourced in discussion with community members on a public Mozilla discussion forum (and seems to have also been informed by discussion with linguists)\footnote{https://discourse.mozilla.org/t/feedback-needed-languages-and-accents-strategy/40352/56}. 

For smaller or marginalised speech communities and/or those in the Global South in particular, this participatory framework of crowdsourcing both language and language policy appears a better strategy for speech and language technology development than relying on large for-profit corporations. Speakers can engage in ``conscious data contribution'' \cite{Vincent2021}, and (within limits) directly shape what kind of language(s) DeepSpeech will support. For some varieties, like Kabyle (a Berber language with 7 million speakers) or Kinyarwanda (a Niger-Congo language with 12 million speakers) this approach also appears successful as they have sizeable validated corpora. However, some varieties (regardless of speaker numbers) have only very small CommonVoice corpora, or corpora which are very unbalanced across varieties  (most notably Arabic, which has a large number of distinct dialects spoken in different regions but is currently only represented by Standard Arabic), as well as age and gender groups. The majority of the contributors to the English CommonVoice corpus, for example, did not provide any information about their accent and only 15\% identified themselves as female. Notably, in the context of existing research on bias in ASR, CommonVoice does not collect information on race or ethnicity, and ``African American English'' is not one of the possible ``native accents''. This lack of documentation makes evaluation of data bias difficult. 

Overall, while crowdsourcing can alleviate some of the data bias issues we see in commercial ASR, especially when done with an explicit focus on accent diversity, many representation issues persist. Recall that \newcite{Meyer2020} show that DeepSpeech (trained on English CommonVoice) produces higher error rates for Indian English speakers than for American English speakers, and \newcite{Martin2020} show that it performs worse for AAE speakers. As has also been discussed in the context of Wikipedia, who contributes to crowdsourced projects depends on many factors such as availability of free time, technical skills, access to digital technology and the culture of the crowdsourced project \cite{hargittaiMindSkillsGap2015,tripodiMsCategorizedGender2021}. Crowdsourcing also places the onus to create data on potentially already marginalised speech communities who might furthermore disagree about how (and if) their language should be represented in these systems (e.g. which accents or writing systems) and how they would like any finished system to be used. 

\subsection{Testing}

In speech and language technologies (and machine learning more broadly), benchmark datasets are used to evaluate the performance of new algorithmic systems \cite{Schlangen2020}. While this focus on benchmarks  has recently become the subject of critique \cite{bowman-dahl-2021-will,denton2020bringing,kochReducedReusedRecycled2021,rajiAIEverythingWhole2021}, they are still central to the way the field defines ``progress''. In the following section we explore how language ideologies shape the well-established academic benchmark corpora TIMIT (English) \citelanguageresource{timit}, Switchboard \citelanguageresource{switchboard} and CallHome (American English) \citelanguageresource{callhome}. 

\subsubsection{Data bias in academic corpora}

 TIMIT (English) \citelanguageresource{timit}, Switchboard \citelanguageresource{switchboard} and CallHome (American English) \citelanguageresource{callhome} are well-established licensed speech corpora which were collected in the late 1980s and early 1990s and are held by the Linguistic Data Consortium. TIMIT (English) was collected by MIT, SRI International and Texas Instruments to be used in speech technology development and acoustic-phonetic research. It features recordings of 630 speakers of 8 ``major dialects of American English'', each reading 10 phonetically rich sentences which have been phonetically transcribed and aligned. Switchboard contains 2,400 two-sided telephone conversations between 543 US American strangers on one of 70 pre-selected topics collected by Texas Instruments. CallHome features 120 unscripted 30-minute telephone conversations between friends or family members (all ``native speakers of English'' who grew up in the United States) and was collected by the Linguistic Data Consortium. 
 
 While all three corpora were carefully designed to capture some regional dialectal variation in US English, they are not balanced across gender groups. Further, most speakers appear to be White, though race is only recorded in the documentation of TIMIT. In the case of TIMIT this is perhaps due to convenience sampling of participants: most of the speakers were employees of Texas Instruments in Dallas which collected the corpus\footnote{https://nvlpubs.nist.gov/nistpubs/Legacy/IR/nistir4930.pdf}. Demographic imbalances are potentially more critical for Switchboard, where only the topic of conversation, not the speech style was constrained and for CallHome where speech styles could also vary widely, and women are over-represented. As noted in \ref{sec:socio101}, this gender imbalance could be indicative of a speech style imbalance. A recent analysis by \newcite{Martin2021} further confirms that Switchboard and TIMIT under-represent AAE.\par

\subsubsection{Evaluation bias \& biased benchmarks}

Systems trained on biased datasets can exhibit predictive bias. But training is not the only context in which harms and biases can be introduced in the development and implementation of a machine learning system. \newcite{SureshGuttag2021} use the term ``evaluation bias'' to describe the bias which occurs when there's a mismatch between the benchmark data used for a particular task and the intended use population. As outlined above, some established benchmarks are unrepresentative of the potential user base of English language ASR, which include second language speakers, speakers of ``non-standard'' regional dialects and ethnolects and speakers who frequently code-switch between several varieties. These benchmarks are also in some ways misaligned to current ASR applications \cite{szymanski-etal-2020-wer}. Today, ASR is widely used to transcribe conversational speech which is notoriously challenging for systems designed to recognise simple commands for virtual agents in human-computer directed speech. 

Particular evaluation strategies can exacerbate this kind of bias \cite{SureshGuttag2021}. Computing an aggregate word error rate across these homogeneous and/or unrepresentative test sets hides predictive bias. If \newcite{Koenecke2020}, for example, had computed word error rate over all speakers, the overall higher than state-of-the-art word error rate would have perhaps been attributed to the conversational nature of the recordings, rather than significant difference by speaker race. As discussed in \ref{ASRbias101}, and as the CORAAL \citelanguageresource{CORAAL} recordings used by \newcite{Koenecke2020} illustrate, race and gender interact in language variation. This is reflective of the concept of intersectionality originating in Black feminist thought \cite{Crenshaw1991}, which recognises that interacting social categories (and axes of oppression) such as race and gender cannot be considered separately. Intersectional evaluation, then, is mindful of these interactions and can capture the differences in life experiences and linguistic behaviours between, for example, Black women and White women, rather than considering either only race or only gender. Within machine learning, this type of approach to evaluation has also been successfully applied in the context of facial analysis \cite{pmlr-v81-buolamwini18a}. 

It is difficult to ascertain how much language ideologies influenced the collection of these licensed corpora in the 1980s and 1990s. At the time, they were created for a relatively narrow purpose (to research speech technologies, particularly in an academic context). It is unlikely that the researchers designing the data collection expected these resources to still be used to benchmark state-of-the-art speech recognition systems thirty years later. While incorporating some regional dialectal variation was clearly a priority, ethnic diversity or the inclusion of African American English wasn't. 

The decision to use these datasets as benchmarks in the 2020s despite these limitations is, however, a choice that constitutes language policy. Just as particular language varieties or datasets are ``selected'' in training, they are also selected in testing. And just as training is shaped by language policy, so is testing. At first glance, Switchboard, TIMIT and CallHome fulfil the primary function of a benchmark: to allow comparison with other systems. Following \newcite{Schlangen2020}'s definition of a benchmark, they should, however, also ``exemplify'' the overall task of interest. A mismatch between benchmark and real-world application is therefore undesirable. More importantly, a mismatch is unexpected, as there is an implied relationship between benchmark and real-life application. The selection of an unrepresentative benchmark is shaped by beliefs about what kind of speech (and by extension, what kind of speakers) speech recognition should (be expected to) work for. Due to the evaluation bias this application of benchmarks produces, these ideologies are then further reinforced. Failure to perform accurately on underrepresented speech not only goes undetected, but, perhaps more troublingly, is not penalised. Of course, the benchmark doesn't have to be representative of all application contexts if we choose to only use it to compare new systems to older systems. But nevertheless, the picture benchmarks provide are always partial and potentially very misleading, especially since they are almost never described in detail in the papers that use them to evaluate \cite{szymanski-etal-2020-wer}.

\section{Towards better practices}

As we tried to highlight in this paper, both the curation and the use of particular speech datasets constitutes a form of language management, itself influenced by beliefs and ideologies surrounding language variation. Given the potentially far-reaching consequences of their decisions, practitioners working with speech datasets could be considered ``language policy arbiters'': individuals who ``[wield] a disproportionate amount of power
in how a policy gets created, interpreted, appropriated, or instantiated
relative to other individuals in the same context'' \cite[100]{Johnson2013}. Who gets to select which data is used in training and testing obviously depends on the broader institutional context. In a commercial context, language policy appears to be primarily driven by (linguistic) markets, and may be decided by business strategists, rather than technologists. But even in commercial contexts, researchers can reflect critically on those policies and, as work in language policy highlights, often have some leeway in the way they implement them \cite{Hornbergeretal2007}. This kind of agentive work is easier in academic research and open-source development. 

 In this final section, we also echo other critical work in machine learning \cite{paulladaDataItsDis2021,Hutchinson_etal2021} and argue that understanding (speech) datasets as increasingly important infrastructure is useful. It allows us to reframe the task of speech technology development from one primarily done by corporations for markets to one done by a wider range of actors for speech communities.

\subsection{Speech technology design as civic design}
A central obstacle to minimising predictive bias in commercial ASR systems appears to be a lack of incentive for corporations to do so. Smaller and more marginalised speech communities are unlikely to be seen as desirable markets by big technology companies, and curating very large datasets could be challenging and relatively expensive. Where proprietary datasets derived from user-data do exist, evaluating data bias is potentially difficult. It’s unlikely that a technology company would be able to document or reliably infer important demographic information (such as accent, age, gender, race) about the speakers whose data is used to create a balanced dataset \cite{Andrus2021}. Curated licensed corpora could be combined to train complex systems (as was done by Microsoft in: \newcite{Microsoft2017}) but since current well-established corpora only represent a small section of all English speakers, new corpora would have to be collected for this purpose. Speech technology companies could, of course, do this themselves, for example by offering payment to users (or crowd-workers) who complete a survey about their demographic background and provide speech recordings of read or naturalistic speech (see Facebook AI’s \newcite{Hazirbas2021} for one of the first attempts at this method). Ultimately, however, this approach would not solve the fundamental issues arising from designing for markets.\par
Alternatively, we could reframe speech technology as a kind of infrastructure and its design as civic design. \newcite[25]{MugarGabriel2020MICD} define “civic design” as an approach to design that “creates the conditions for a plurality of voices and interests to be represented, accounted for, and involved in shaping the outputs and effects of public life.” Civic design is design for and with publics, rather than markets \cite[53]{MugarGabriel2020MICD}. The notion of a “public” as a collective of people which emerges through discursive circulation of shared interests with the purpose of influencing decision-making \cite[66]{MugarGabriel2020MICD}, has also been taken up in the analysis of groups of language users \cite{Muehlmann2014,Gal1995}. \par
Some linguistic publics intersect with the public of a nation (state), such as the Icelandic-speaking public or the Estonian-speaking public. In those cases, a (national) government (a traditional actor in language policy) shares the public’s interest in the development of speech technologies which it understands as a type of infrastructure. It can steer (and pay for) corpus development. The governments of Iceland and Estonia have both overseen design and development of open-source speech and language technology resources (corpora and models) by private and public partners \cite{Nikulasdottir2020}. Similarly, the Welsh government has prioritised speech and language technology development and is working with universities and private sector businesses to deliver it \cite{WelshSLT2018,WelshProgressReport2020}. \par
A civic design approach can also be useful for other kinds of diverse linguistic publics which do not necessarily form a “viable market”. As digital devices are becoming crucial gateways to accessing public services, jobs, and media and predictive bias could exclude many people from using them. The public sector (including but not limited to governments) is potentially well positioned to drive the development of speech technologies as accessibility tools. Civic design as something that is done by a public for a public also has the potential to resolve some of the current issues with crowd-sourcing speech datasets. By carefully (and meaningfully) engaging speakers, not just as anonymous data sources, but as co-designers who can shape the technology development process, following, for example, principles of design justice \cite{2020Design}, technology developers (private or public) would likely be able to create more representative and ultimately more useful technologies, and move away from colonial frames inherent in many drives to “spread language technologies” \cite{bird-2020-decolonising}. With the proliferation of open-source speech technology toolkits and cheap(er) cloud computing, some publics may be able to build or modify these technologies without much or any support from governments of corporations (e.g. Masakhane\footnote{https://www.masakhane.io/home}, \newcite{khandelwal20_interspeech}).
\newcite{MugarGabriel2020MICD} also emphasise that, in their vision, the aim of civic design by and for publics is care rather than innovation, and space for meaningful interaction between people, rather than efficiency. These values run counter to the ethos currently driving commercial technology development, but they are excellent principles in the context of technology designed fundamentally to facilitate communication.
\subsection{Speech datasets as infrastructure}

Whether speech technologies are approached from a perspective of civic design or not, speech datasets are, like all datasets in machine learning, infrastructure. As \newcite{Hutchinson_etal2021} point out, curation and maintenance of this infrastructure is undervalued in the machine learning community and as a result, datasets are often poorly documented and precariously stored. Fundamentally, careful curation (following a civic design model or any other model) and good documentation of speech corpora is tractable, due to the comparatively smaller size of datasets compared to, for example, large language models \cite{Bender2020}. Documentation is essential in mitigating (or even simply anticipating) predictive bias. Speech datasets (like other language datasets) need not be static but rather, like physical infrastructure, require maintenance and updating. As language (both in use and form) continuously changes, static datasets will deprecate over time and an approach in which practitioners can add or remove data from training sets in deployment may be more useful (assuming any changes are documented).

\section{Conclusion}


Predictive bias in speech recognition technologies is an increasingly important problem as speech recognition systems get embedded into complex algorithmic systems, with harms disproportionately falling on already marginalised speech communities. We believe that language policy is a lens that can empower technologists to mitigate data bias and recognise potential harms of biased technologies. We want to encourage practitioners to adopt this reflexive approach to better understand how language ideologies affect speech technologies and their users, and to use this understanding to build better speech technologies.

\section{Acknowledgements}

This work was supported in part by the UKRI Centre for Doctoral Training in Natural Language Processing, funded by the UKRI (grant EP/S022481/1) and the University of Edinburgh, School of Informatics and School of Philosophy, Psychology \& Language Sciences. This work was also partially made possible by an ESRC studentship (grant ES/P000681/1). We'd also like to thank Catherine Lai, Lauren Hall-Lew, Bonnie Webber, John Joseph and Joseph Gafaranga for helpful discussion. 

\section{Bibliographical References}\label{reference}

\bibliographystyle{lrec2022-bib}
\bibliography{LREC.bib} 

\section{Language Resource References}
\label{lr:ref}
\bibliographystylelanguageresource{lrec2022-bib}
\bibliographylanguageresource{languageresource}

\end{document}